\documentclass{article}[10pt]
\usepackage[margin=1in]{geometry}
\usepackage[utf8]{inputenc}
\usepackage{graphicx}
\usepackage{graphics}
\usepackage{amsmath}
\usepackage{amsfonts}
\usepackage{subcaption}\usepackage{rotating}
\usepackage[
backend=biber,
style=numeric,
maxcitenames=1
]{biblatex}
\usepackage{booktabs}
\usepackage{dsfont}
\usepackage{wrapfig}
\usepackage{float}
\usepackage{multirow}
\makeatletter

\newenvironment{figurehere}
  {\def\@captype{figure}}
  {}
\makeatother
\usepackage{caption}
\usepackage{multicol}
\usepackage{authblk}

\usepackage{titlesec}
\titleformat{\section}
  {\normalfont\fontsize{10}{15}\bfseries}{\thesection}{1em}{}
\titleformat{\subsection}
  {\normalfont\fontsize{10}{15}\bfseries}{\thesubsection}{1em}{}

\title{An Encoding Framework for Binarized Images using HyperDimensional Computing}
\author[1,*]{Laura Smets}
\author[1]{Werner Van Leekwijck}
\author[1]{Ing Jyh Tsang}
\author[1]{Steven Latr\'e}
\affil[1]{University of Antwerp - imec, IDLab - Department of Computer Science, Sint-Pietersvliet 7, 2000 Antwerp, Belgium}
\affil[*]{Corresponding author, E-mail: Laura.Smets@uantwerpen.be}
\date{\vspace{-5ex}}

\begin{document}

\maketitle

\section*{Abstract}

Hyperdimensional Computing (HDC) is a brain-inspired and light-weight machine learning method. It has received significant attention in the literature as a candidate to be applied in the wearable internet of things, near-sensor artificial intelligence applications and on-device processing. HDC is computationally less complex than traditional deep learning algorithms and typically achieves moderate to good classification performance. A key aspect that determines the performance of HDC is the encoding of the input data to the hyperdimensional (HD) space. This article proposes a novel light-weight approach relying only on native HD arithmetic vector operations to encode binarized images that preserves similarity of patterns at nearby locations by using point of interest selection and \textit{local linear mapping}. The method reaches an accuracy of 97.35\% on the test set for the MNIST data set and 84.12\% for the Fashion-MNIST data set. These results outperform other studies using baseline HDC with different encoding approaches and are on par with more complex hybrid HDC models. The proposed encoding approach also demonstrates a higher robustness to noise and blur compared to the baseline encoding.

\section*{Keywords}

Hyperdimensional Computing - Vector Symbolic Architectures - Image Encoding - Image Classification - Handwritten Digit Recognition

\begin{multicols}{2}

\section{Introduction} \label{sec:1}

Because of the rising interest in the wearable internet of things (IoT), near-sensor artificial intelligence (AI) applications and on-device processing, there is considerable need for energy-efficient algorithms. Hyperdimensional computing (HDC) has been proposed in the literature as a brain-inspired, light-weight and energy-efficient method because it has the advantages of few data requirement \cite{Rahimi2019}, robustness to noise \cite{Kanerva2009,Widdows2015,Rahimi2019}, low latency \cite{Rahimi2019} and fast processing \cite{Rahimi2019}. HDC maps input data to a hyperdimensional (HD) space in which information is distributed across thousands of vector elements, mimicking the large number of neurons that store information in our brains. Since HDC uses simple HD arithmetic operations, it is computationally less complex than traditional deep learning (DL). HDC has already been used in several applications, such as speech recognition \cite{Imani2017}, human activity recognition \cite{Kim2018}, hand gesture recognition \cite{Rahimi2016a, Moin2021, Zhou2021}, text classification \cite{Rachkovskij2007}, classification of medical images \cite{Kleyko2017a, Watkinson2021}, character recognition \cite{Manabat2019}, robotics \cite{Neubert2019}, and time series classification \cite{Schlegel2022}. \\

\noindent A crucial aspect that determines the performance of HDC is the encoding of the input data to the HD space, which highly depends on the type of input data. To date, studies have clearly defined how text data \cite{Kleyko2022a}, numeric data \cite{Imani2017,Kim2018} and time-series data \cite{Rahimi2016a} can be encoded in a simple way using the HD arithmetic operations. However, what is still missing in the literature is a uniform framework to encode (binarized) images. Therefore, this article aims to propose a novel light-weight HD approach to encode binarized images relying only on native HD arithmetic vector operations. In this aspect, the current article brings forward the following novelties:
\begin{enumerate}
    \item \textit{Local linear mapping} is introduced as a novel mapping method for numeric data, whereby nearby numerical values are represented by similar HD vectors, and all other values by orthogonal HD vectors. In addition, we demonstrate its application for encoding positions in 2D images;
    \item A parameterized framework for encoding binary images into HD vectors is defined which uses point of interest (POI) selection as a local feature extraction method and unifies existing approaches for native HD encoding of images;
    \item The proposed framework is applied on benchmark data sets, reaching 97.35\% classification accuracy on MNIST and 84.12\% accuracy on Fashion-MNIST. \\
\end{enumerate}

\noindent The remainder of the article is structured as follows. In the next section, the HDC model for classification is described. Afterwards, section 3 defines the \textit{local linear mapping} for numeric data and illustrates its application to 2D position encoding. Section 4 provides an overview of encoding approaches for binarized images found in the literature and introduces our parameterized unified framework. The fifth section describes the performed experiments to test the proposed encoding framework, whereafter the results are presented and discussed in the sixth and seventh section, respectively. The last section presents the conclusions of the article.

\section{Hyperdimensional Computing} \label{sec:2}

HDC is a mathematical framework using HD vectors (i.e., vectors with very high dimension typically up to ten thousands, also called hypervectors (HVs)) and simple HD arithmetic vector operations to represent data. The focus of this article is on dense binary HVs (i.e., the elements are 0 or 1 with an equal probability of occurrence of both values) of dimension 10,000 \cite{Smets2023}. The analysis of data relies on the similarity between HVs which is calculated using the normalized Hamming distance between two binary HVs $\textbf{v}_{1}$ and $\textbf{v}_{2}$\footnote{\normalsize A list of used symbols can be found in the appendix (Table \ref{tab:symbols}).}:
\begin{equation} \label{eq:similarity}
    s(\textbf{v}_{1},\textbf{v}_{2}) = 1 - \frac{h(\textbf{v}_{1},\textbf{v}_{2})}{D}
\end{equation}
with $s$ the similarity between $\textbf{v}_{1}$ and $\textbf{v}_{2}$, $D$ the dimensionality and $h$ the Hamming distance between $\textbf{v}_{1}$ and $\textbf{v}_{2}$:
\begin{equation} \label{eq:hamming}
    h(\textbf{v}_{1},\textbf{v}_{2}) = \sum_{d = 1}^{D} (\textbf{v}_{1}[d] \textrm{ XOR } \textbf{v}_{2}[d]).
\end{equation}
The HD arithmetic vector operations include: \\
\textbf{(a) bundling} $\oplus$: $\mathcal{B} \times \mathcal{H} \to \mathcal{B}$: $(\textbf{B},\textbf{v}) \to \textbf{B}+\textbf{v}$ where $\mathcal{B} = \mathbb{N}^{D}$ and $\mathcal{H} = \{0,1\}^{D}$ (i.e., element-wise addition) after which the bundle $\textbf{B}$ is binarized into the HV $\textbf{v}$ with the majority rule $[.]: \mathcal{B} \to \mathcal{H}$: $\textbf{B} \to \textbf{v}$ according to:
\begin{equation} \label{eq:majority}
    \textbf{v}[d] = [\textbf{B}[d]] =
    \begin{cases}
        1         & \text{if} \, \textbf{B}[d] > \frac{n}{2} \\
        0         & \text{if} \, \textbf{B}[d] < \frac{n}{2} \\
        rand(0,1) & \text{if} \, \textbf{B}[d] = \frac{n}{2}
    \end{cases}
\end{equation}
with $n$ the number of HVs bundled in $\textbf{B}$ and $rand(0,1)$ means that the component $\textbf{v}[d]$ is randomly assigned to 0 or 1 in the presence of ties; \\
\textbf{(b) binding} $\otimes$: $\mathcal{H} \times \mathcal{H} \to \mathcal{H}$: $(\textbf{v}_{1},\textbf{v}_{2}) \to \textbf{v}_{1} \textrm{ XOR } \textbf{v}_{2}$; and \\
\textbf{(c) permutation} $\rho$: $\mathcal{H} \to \mathcal{H}$ (e.g., cyclic shift in binary HDC). \\

\noindent Figure \ref{fig:overview} gives a schematic overview of the framework of HDC in which two main building blocks can be distinguished: an encoder and a classifier. The encoder is responsible for mapping the input to an HV. Typically, it maps each input value of a sample to an atomic HV that is stored in (continuous) item memories ((C)IM). This procedure is called mapping and will be explained in section \ref{sec:3}. Then, different atomic HVs are combined to obtain one sample HV for each input. \\

\begin{figurehere}
\centering
\includegraphics[width=\linewidth]{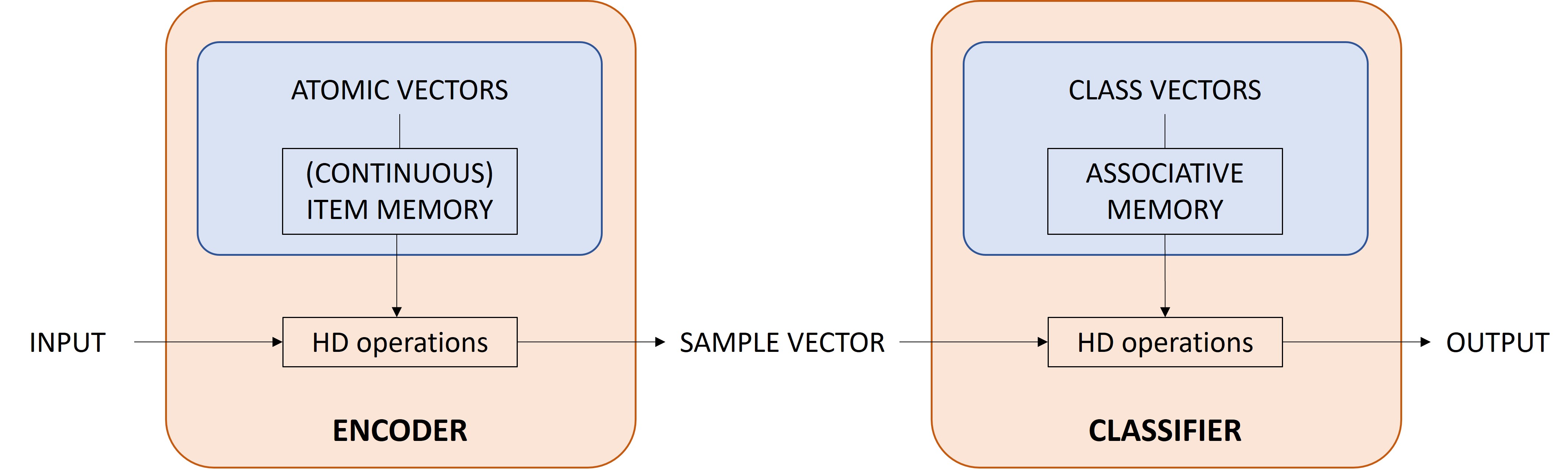}
\caption{Schematic overview of the HDC framework in which two main building blocks can be distinguished: an encoder and a classifier.}
\label{fig:overview}
\end{figurehere}

\noindent Commonly, an input sample $f$ having $n$ features is encoded with the so-called record-based encoding \cite{Imani2018,Kussul1991b,Rachkovskij1990} as Figure \ref{fig:spatial}: Each feature ($j = 1...n$) is assigned a random HV to represent the feature ID which is stored in an IM. Feature values are translated in HVs with a CIM that is created with \textit{linear mapping} (see section \ref{sec:3.2}) \cite{Kleyko2018b,Rahimi2016a}. Next, each feature ID HV $\textbf{v}_{j}$ is bound with the HV representing its value $\textbf{v}_{f[j]}$. Finally, these ID-value bound pairs of all features are bundled together to form the sample bundle $\textbf{S}$ by initializing
\begin{equation} \label{eq:samplesa}
    \textbf{B}_{0} = \{0\}^D
\end{equation}
and bundling each bound pair $\textbf{v}_{f[j]} \otimes \textbf{v}_{j}$ one at a time:
\begin{equation} \label{eq:samplesb}
    \textbf{B}_{j} = \textbf{B}_{j-1} \oplus (\textbf{v}_{f[j]} \otimes \textbf{v}_{j}).
\end{equation}
The sample bundle \textbf{S} is then simply:
\begin{equation} \label{eq:samplesc}
    \textbf{S} = \textbf{B}_{n}.
\end{equation}
For notation purposes, this iterative bundling (Equation \ref{eq:samplesa}-\ref{eq:samplesc}) will be written in short as:
\begin{equation} \label{eq:samples}
    \textbf{S} = \bigoplus_{j=1}^{n} (\textbf{v}_{f[j]} \otimes \textbf{v}_{j})
\end{equation}
Finally, the sample bundle is binarized into the HV $\textbf{s} = [\textbf{S}]$ with the majority rule (Equation \ref{eq:majority}). \\

\begin{figurehere}
\centering
\includegraphics[width=\linewidth]{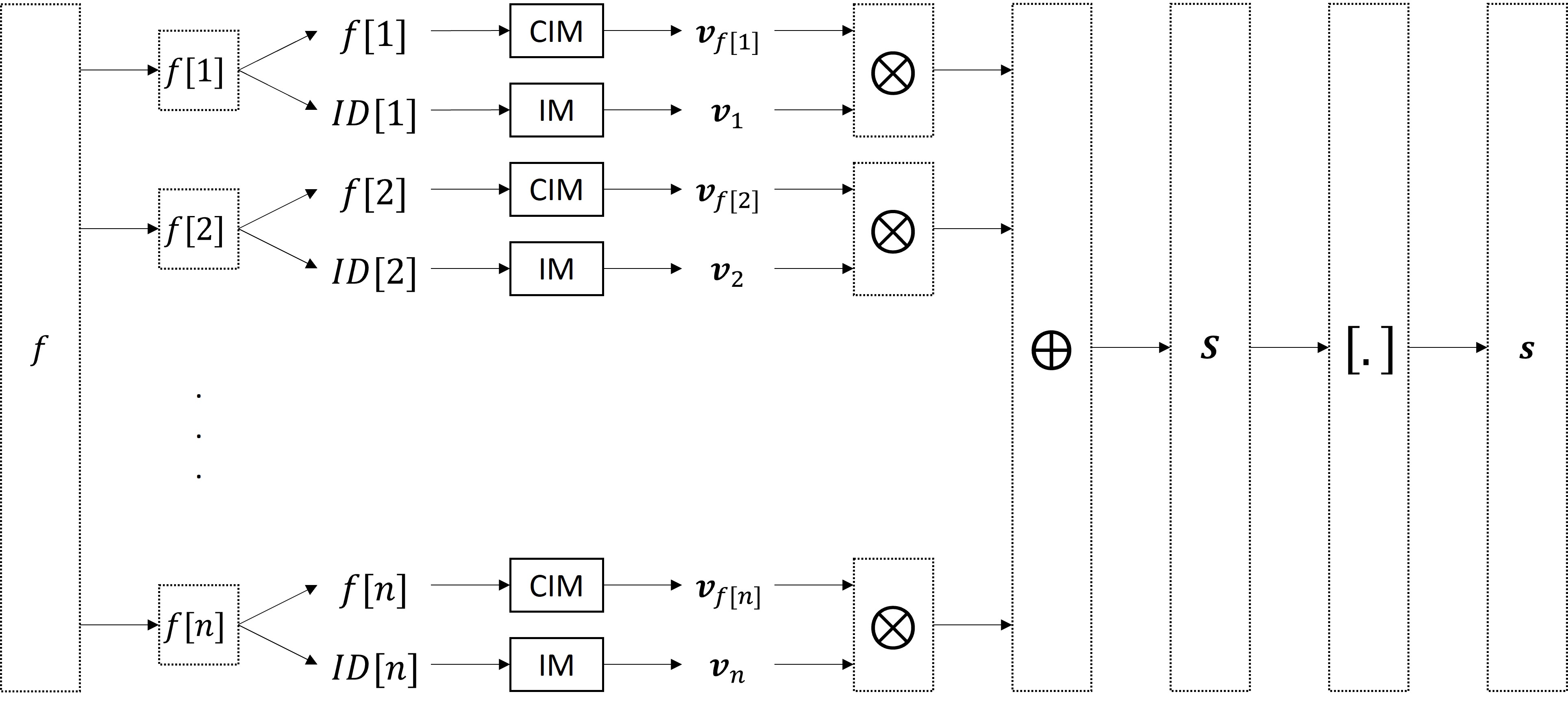}
\caption{Schematic overview of the record-based encoding \cite{Imani2018,Kussul1991b,Rachkovskij1990} with an IM for the feature IDs and a CIM for the feature values.}
\label{fig:spatial}
\end{figurehere}

\noindent As the second main building block, the classifier has two modes of operation: (1) during training, the sample HVs and associated class labels are used to produce class prototypes; and (2) during inference, a sample HV is compared with each of the class prototypes and predicts the corresponding class label by selecting the class with highest similarity. Different variants of training methods exist, as reported in our previous work \cite{Smets2023}. \\

\noindent Since the encoder is a crucial part of the system and a uniform framework to encode (binarized) images is still lacking in the literature, we propose a novel encoding framework (section \ref{sec:4}).

\section{Data mapping techniques} \label{sec:3}

\subsection{Orthogonal mapping} \label{sec:3.1}

Orthogonal mapping assigns a randomly chosen atomic HV to each possible value present in the data. These random HVs are pseudo-orthogonal due to the high dimensionality which converges to exact orthogonality with increasing dimensionality \cite{Kleyko2022a}. This type of mapping is suitable for nominal data where each value is independent from other values. 

\subsection{\textit{Linear mapping}} \label{sec:3.2}

In the case of ordinal or discrete data, there is a natural ordering of levels or values such that closer levels should be mapped to more similar HVs than levels further apart and similarity-preserving HVs are preferred for this type of data. Therefore, \textit{linear mapping} of levels to atomic HVs is applied \cite{Kleyko2018b,Rahimi2016a}. Namely, the lowest level is assigned a random atomic HV, whereafter each level's atomic HV is obtained by flipping $\frac{D/2}{L-1}$ bits in the atomic HV of the previous level, where L is the number of levels (without flipping a bit that has already been flipped before). Similarly, continuous data can be mapped to HVs with \textit{linear mapping} after being quantized into a predefined number of discrete levels. \\

\noindent As an example, Figure \ref{fig:linearmapping} illustrates the application of \textit{linear mapping} for a feature with discrete values ranging from -100 to 100 with steps of 10 and thus 21 levels. It shows the similarity of values to the lowest level (feature value = $-100$) that decreases linearly up until orthogonality (similarity = 0.5) and the similarity of values to the feature value equal to $-30$ that decreases linearly for smaller and larger feature values.

\begin{figurehere}
\centering
\includegraphics[width=.5\textwidth]{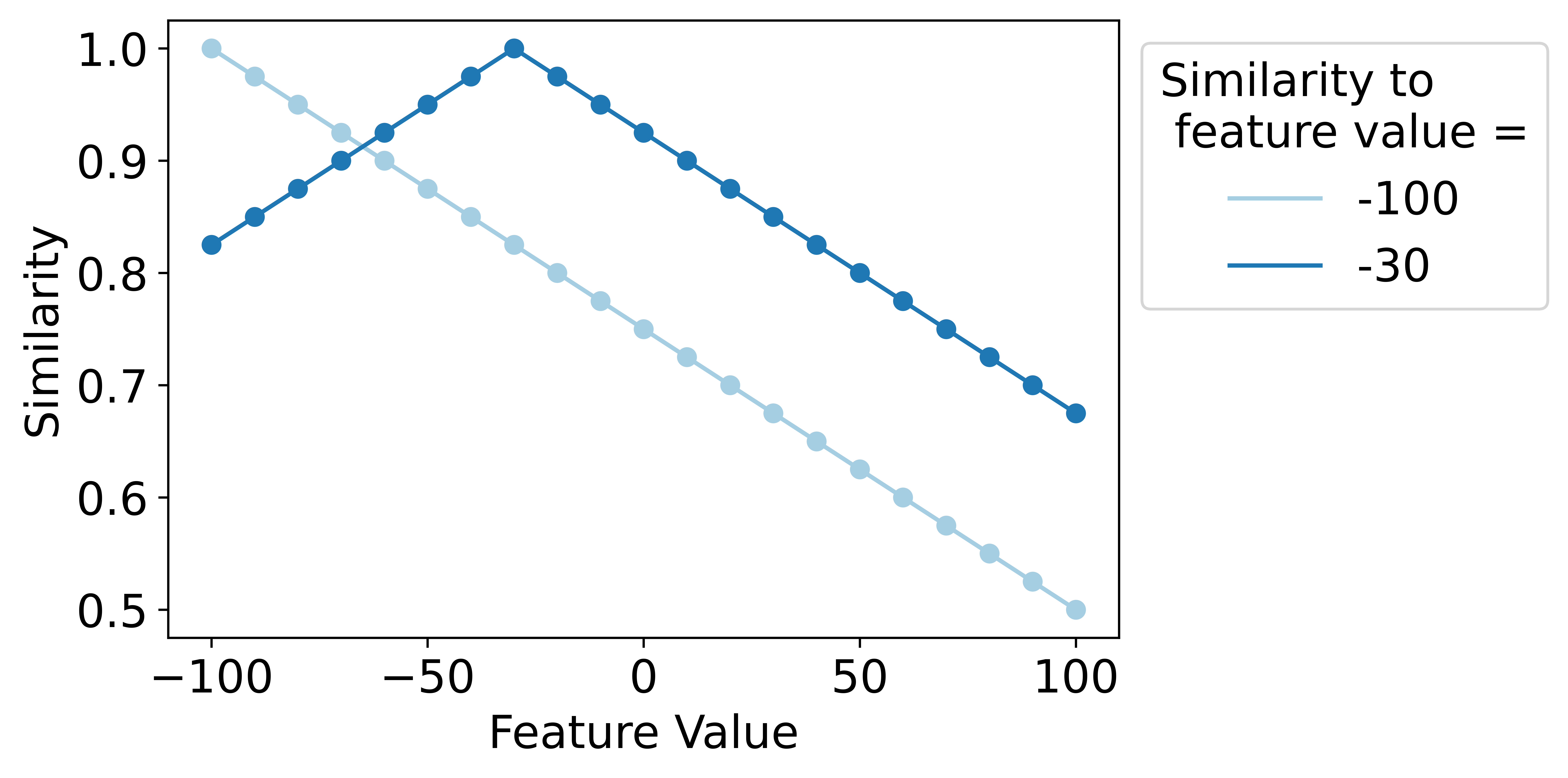}
\caption{Example of \textit{linear mapping} \cite{Kleyko2018b,Rahimi2016a} for a feature with discrete values ranging from -100 to 100 with steps of 10 and thus 21 levels. The similarity of each feature value's level hypervector to the lowest level hypervector (feature value $= -100$) and to the hypervector for the feature value of $-30$ is shown.}
\label{fig:linearmapping}
\end{figurehere}

\subsection{\textit{Local linear mapping}} \label{sec:3.3}

Encoding numeric data with original \textit{linear mapping} results in small differences between the HVs of two adjacent values when working with a relatively large number of levels, and even values that are far apart are always somewhat similar ($s > 0.5$). Therefore, we introduce \textit{local linear mapping} which splits the range of values in $S$ splits such that a smaller number of HVs (i.e., $\frac{L-1}{S}+1$ HVs) is present in each split to which \textit{linear mapping} can be applied. As such, the upper edge vector of a previous split is used to apply \textit{linear mapping} in the following split. Consequently, two adjacent values within one split will have a larger difference in HVs (i.e., $\frac{D/2}{((L-1)/S)}$ different bits) compared to when applying original \textit{linear mapping} to the whole range of values. Additionally, an HV will be similar to HVs within a certain range from the considered HV and dissimilar, thus approximately orthogonal, to all HVs further away from the considered HV (i.e., outside that certain range). As a result, small differences in values are emphasized and large differences are ignored. Note that \textit{local linear mapping} with 1 split or $L$ splits correspond to the original \textit{linear mapping} and orthogonal mapping, respectively. \\

\noindent Figure \ref{fig:disclinearmapping} illustrates the concept of the proposed \textit{local linear mapping} with 4 splits and thus 6 vectors in one split since there are 21 levels (i.e., $\frac{21-1}{4}+1$). In each of the four splits (e.g., between the edge vectors for values -100 and -50), original \textit{linear mapping} is applied. Two adjacent values within one split will be highly similar; an HV will be similar to vectors at nearby positions to the left and right; an HV is orthogonal to vectors further to the left and right. \\

\begin{figurehere}
\centering
\includegraphics[width=.5\textwidth]{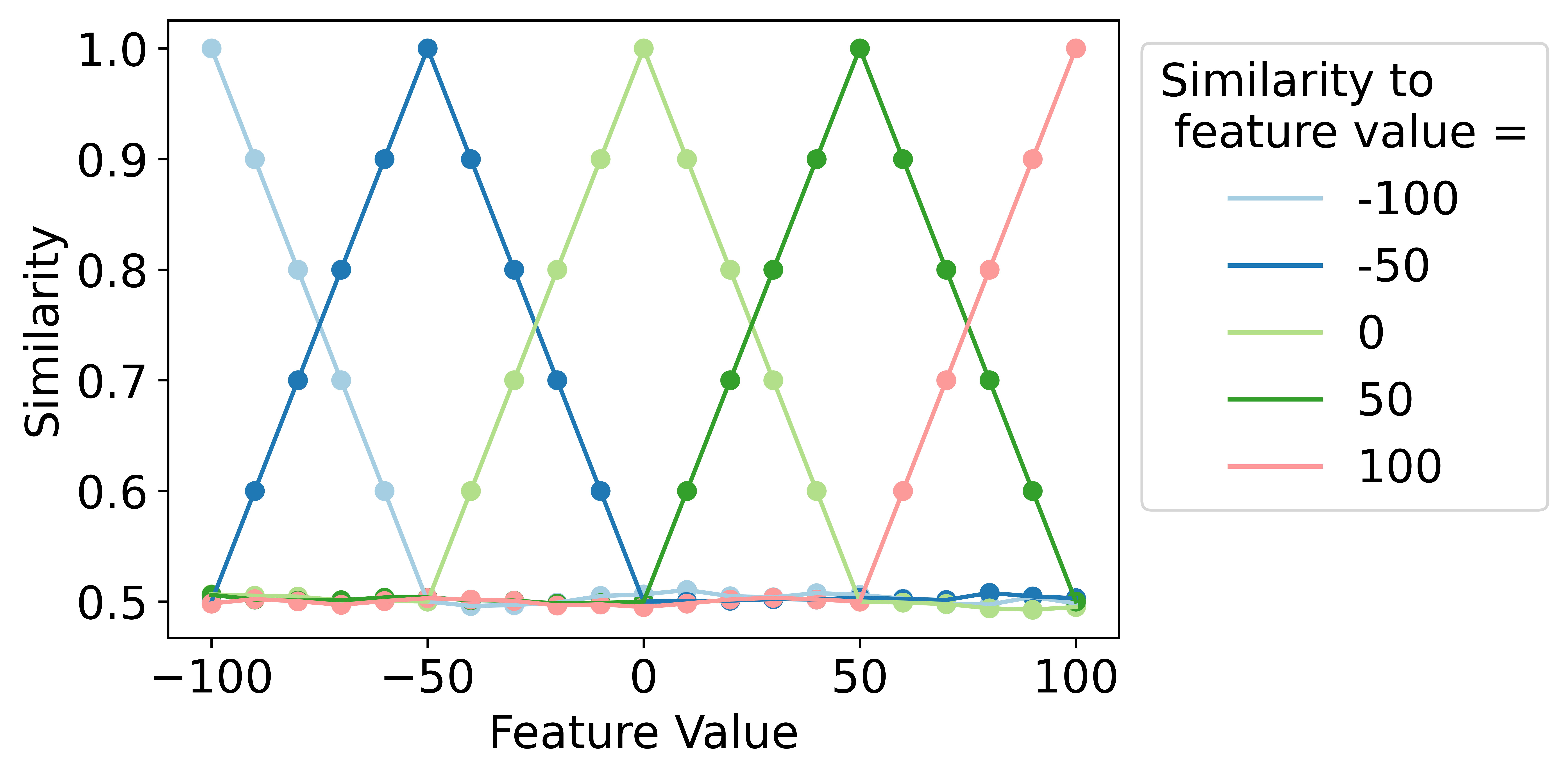}
\caption{Example of \textit{local linear mapping} with 4 splits for a feature with discrete values ranging from -100 to 100 with steps of 10 and thus 21 levels and $\frac{21-1}{4}+1=6$ levels in one split. The similarity of each feature value's level hypervector to each edge hypervector (feature value = -100, -50, 0, 50 and 100) is shown.}
\label{fig:disclinearmapping}
\end{figurehere}

\begin{figure*}[!hbt]
\begin{subfigure}{.5\linewidth}
    \centering
    \includegraphics[scale=.5]{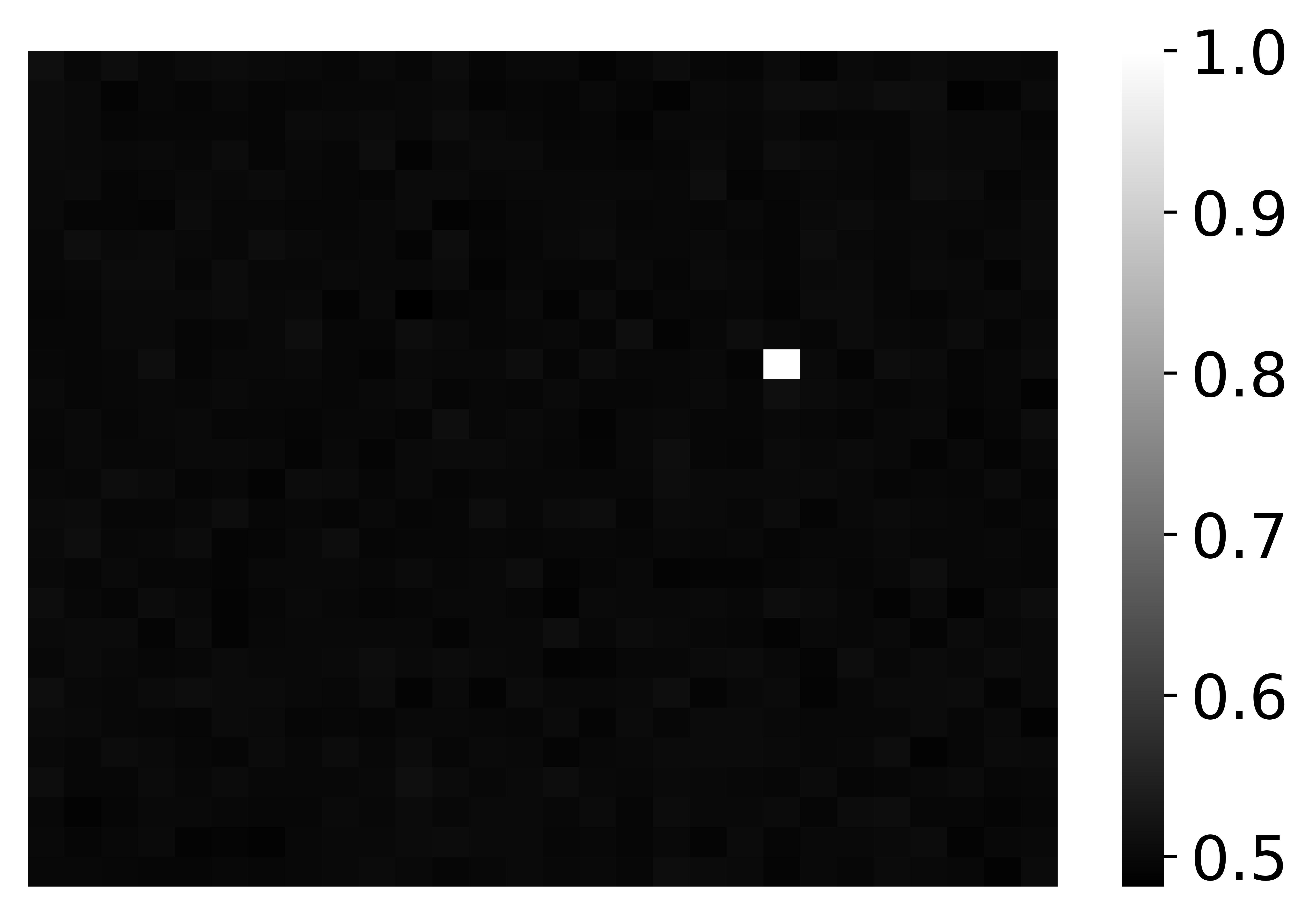}
    \caption{Orthogonal mapping.}
    \label{fig:orthogonal}
\end{subfigure}
\begin{subfigure}{.5\linewidth}
    \centering
    \includegraphics[scale=.5]{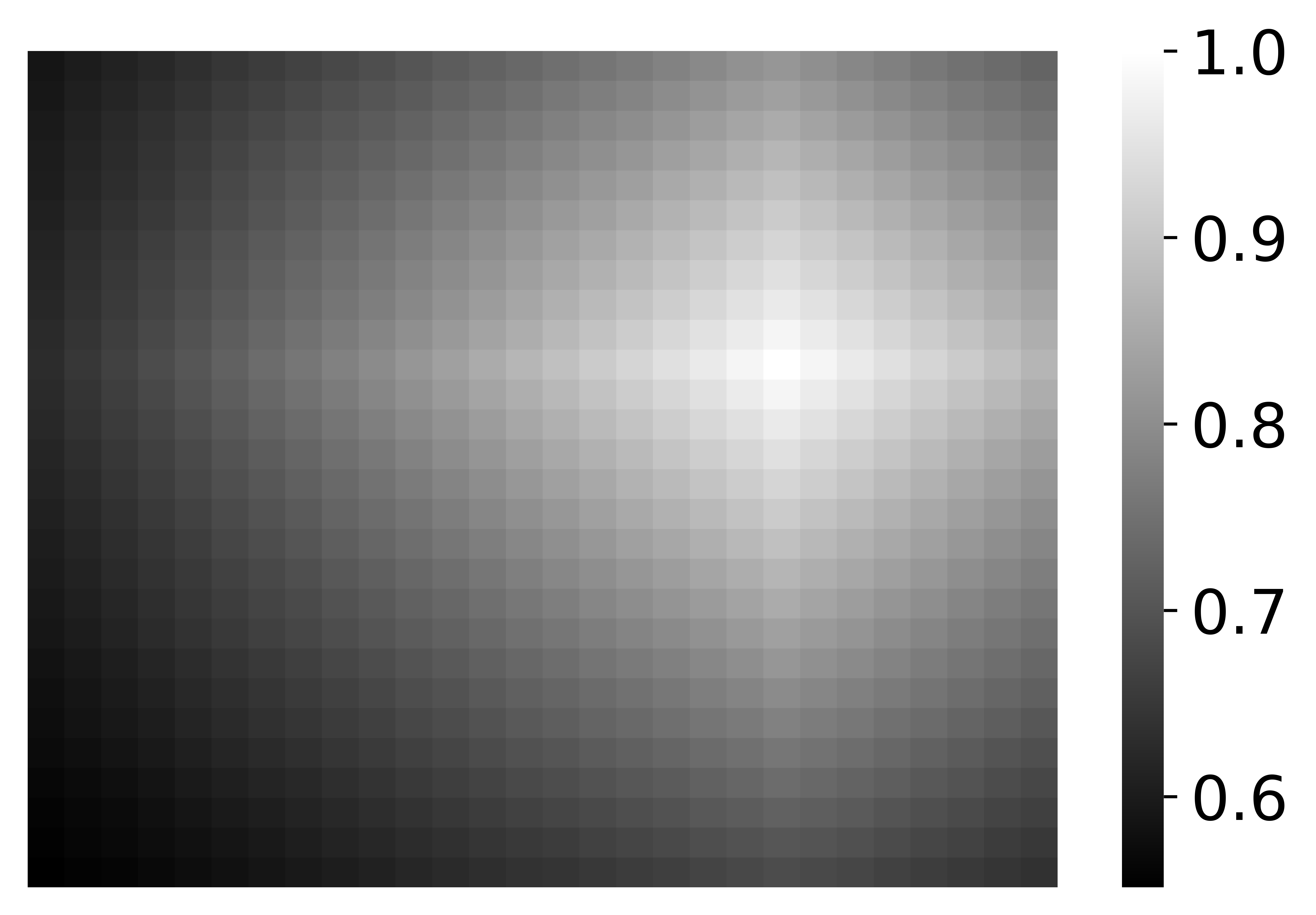}
    \caption{\textit{Linear mapping} \cite{Kleyko2018b,Rahimi2016a}.}
    \label{fig:nosplits}
\end{subfigure}
\vskip \baselineskip
\vspace{-10pt}
\begin{subfigure}{.5\linewidth}
    \centering
    \includegraphics[scale=.5]{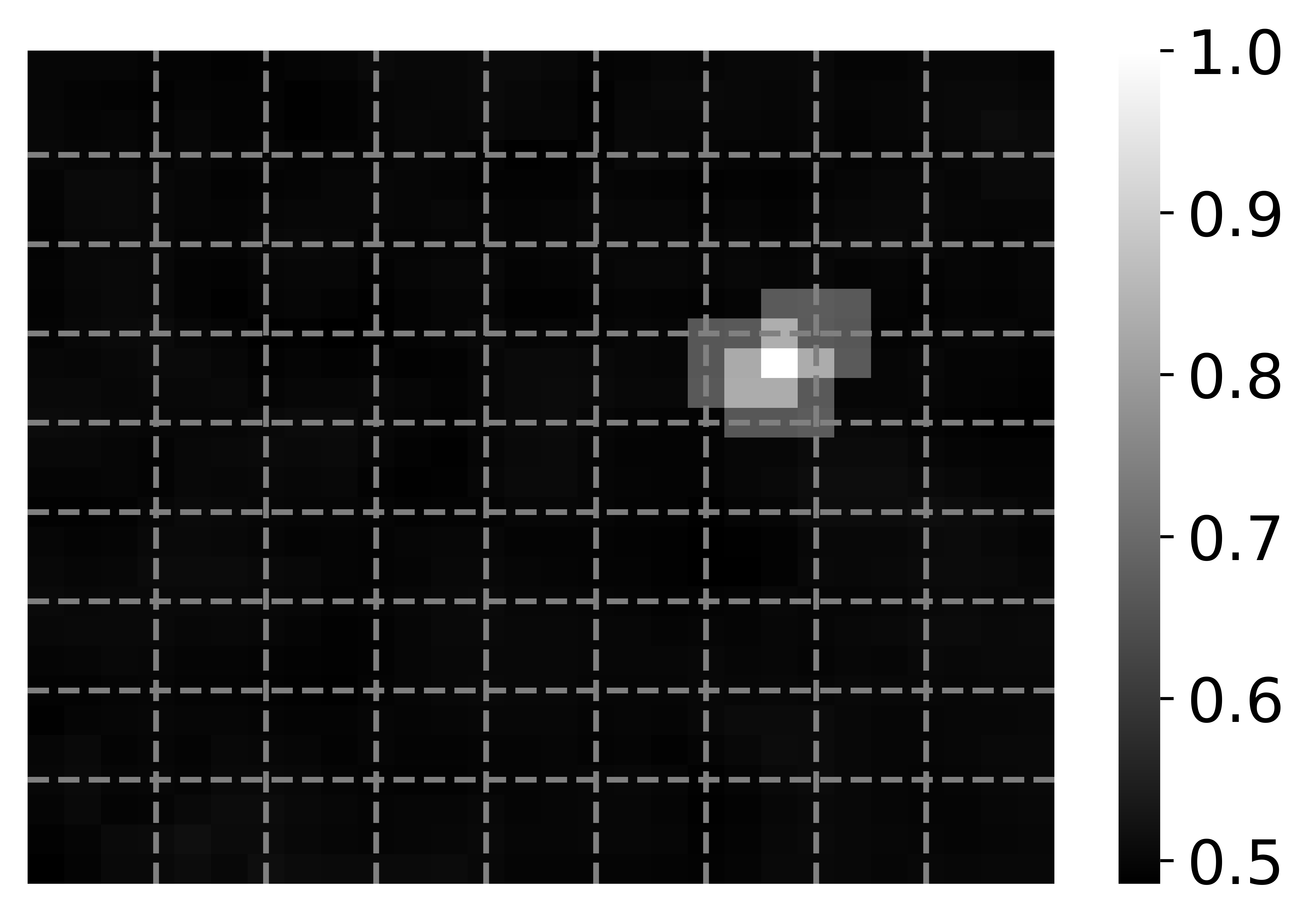}
    \caption{Concatenation \cite{Neubert2021a, Rachkovskij2005a}.}
    \label{fig:neubert}
\end{subfigure}
\begin{subfigure}{.5\linewidth}
    \centering
    \includegraphics[scale=.5]{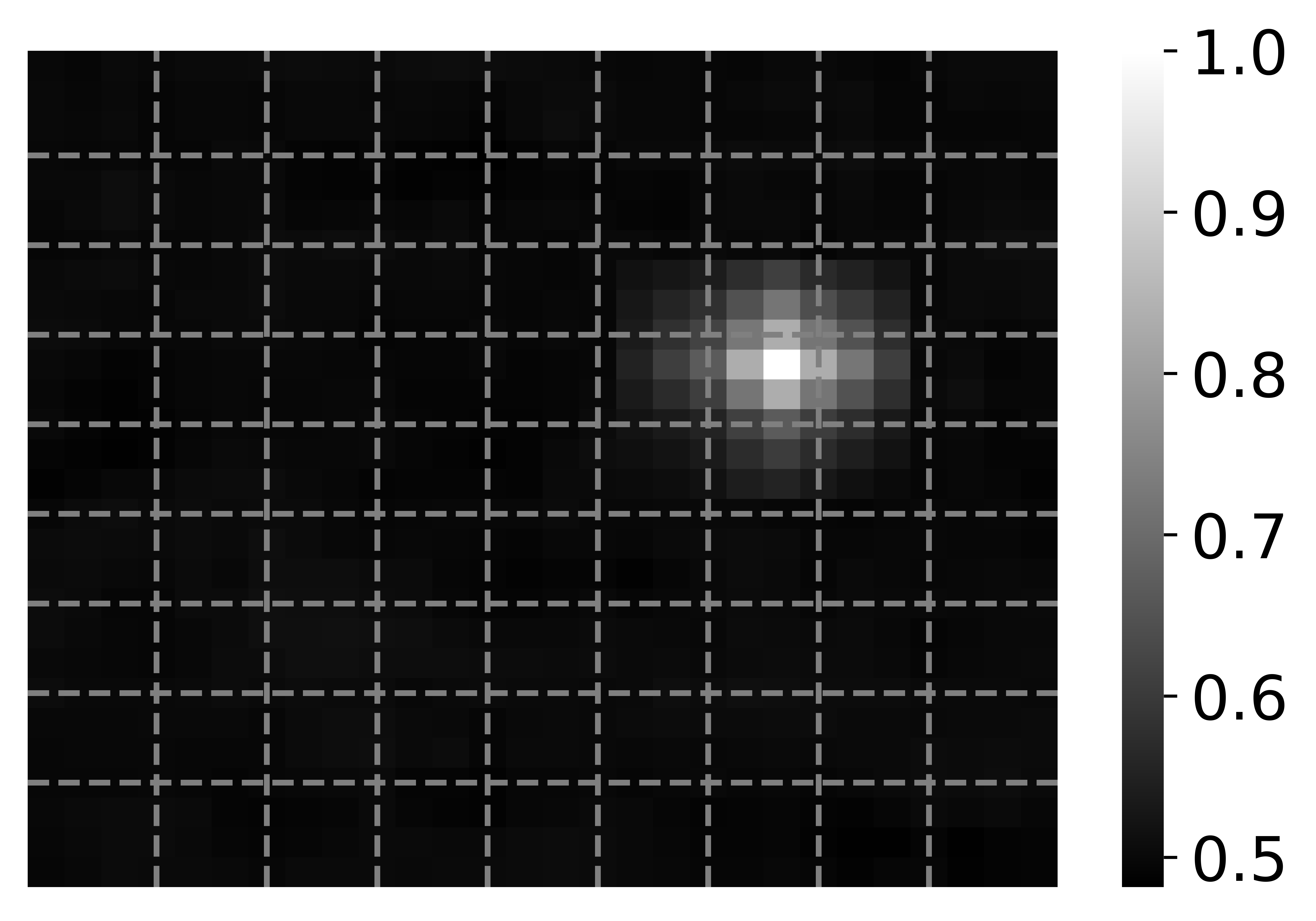}
    \caption{\textit{Local linear mapping}.}
    \label{fig:9splits}
\end{subfigure}
\caption{Similarity of all pixel's position vector to position vector of pixel at location (21,11) for an image of size 28x28 that are encoded as $\textbf{v}_{x} \otimes \textbf{v}_{y}$ of which the x and y positions are mapped to vectors with (a) orthogonal mapping, (b) \textit{linear mapping} \cite{Kleyko2018b,Rahimi2016a}, (c) the concatenation approach of \textcite{Rachkovskij2005a, Neubert2021a} using 10 edge vectors for each axis (dotted lines) and (d) our proposed approach of \textit{local linear mapping} with 9 splits and thus 10 edge vectors (dotted lines).}
\label{fig:mappings}
\end{figure*}

\noindent \textit{Local linear mapping} has some resemblance to a technique introduced by \textcite{Rachkovskij2005a, Neubert2021a} for encoding position in images, which concatenates orthogonal edge vectors to obtain the position vectors within one split. The ratio of concatenation depends on the distance of the considered pixel to both edge vectors. However, the decrease in similarity for pixels further away from the considered pixels is not as gradual as with the proposed \textit{local linear mapping}. This is shown in Figure \ref{fig:mappings} which illustrates the difference in similarity between all pixels' position HV and the position HV of pixel at location (21,11) for an image of size 28x28. The position HVs are all encoded as $\textbf{v}_{x} \otimes \textbf{v}_{y}$ of which the x and y positions are mapped to vectors $\textbf{v}_{x}$ and $\textbf{v}_{y}$ using the different types of mapping: (1) orthogonal mapping, (2) \textit{linear mapping} \cite{Kleyko2018b,Rahimi2016a}, (3) the concatenation approach of \textcite{Rachkovskij2005a, Neubert2021a} using 10 edge vectors and (4) our proposed \textit{local linear mapping} using 9 splits and thus also 10 edge vectors. We believe that the decrease in similarity for \textit{local linear mapping} in Figure \ref{fig:9splits} is more intuitive than for the concatenation approach \cite{Neubert2021a, Rachkovskij2005a} (Figure \ref{fig:neubert}). Furthermore, \textit{local linear mapping} builds further on the concept of \textit{linear mapping} which is commonly used in HDC encoding approaches. In this aspect, it is also similar in concept to float code, which makes the similarity decay local but builds further on thermometer code \cite{Rachkovskij2005a, Frady2021}. \\

\section{Encoding techniques for binary images} \label{sec:4}

\subsection{Related Work} \label{sec:4.1}

Several ways to encode binarized images with HDC have been proposed in the literature and can be divided into two main categories: (1) native HDC, i.e., end-to-end use of native HD vector operations (from raw pixel to output) and (2) hybrid HDC, i.e., external feature extraction methods are used in combination with HDC. Table \ref{tab:literature} gives an overview of the different encoding approaches which are discussed in the following section. \\

\begin{table*}[!hbt]
\centering
\caption{Summary of the already proposed approaches for the encoding of binarized images. The table includes a short description of the type of encoding, the references where the encoding is discussed and the formula to obtain the encoded HV of the image. The symbols used in this table are listed in the appendix (Table \ref{tab:symbols}).}
\label{tab:literature}
\begin{tabular}{@{}rrrrrl@{}}
\hline
\multicolumn{4}{c}{\textbf{Description}} & \textbf{References} & \textbf{Encoded image $\textbf{v}_{I}$}\\
\hline
Native & Orthogonal & Permutation & 1D & \cite{Hassan2022, Kleyko2016, Kleyko2017b, Manabat2019} &  $[\bigoplus_{x=1}^{w*h} (\rho^{p[x]} \textbf{v}_{x})]$ \\
           &                          &             & 2D & \cite{Kleyko2020, Kussul2006, Mitrokhin2019, Rachkovskij2022b} & $[\bigoplus_{x=1}^{w} \bigoplus_{y=1}^{h} (\rho_{X}^{x} \rho_{Y}^{y} \textbf{v}_{I_{bin}[x,y]})]$ \\
           &                          & Binding     & 1D & \cite{Bosch2022, Duan2022a, Ma2022, Ma2021, Watkinson2021, Yang2017b} & $[\bigoplus_{x=1}^{w*h} (\textbf{v}_{x} \otimes \textbf{v}_{p[x]})]$ \\
           &                          &             & 2D & \cite{Kelly2013} & $[\bigoplus_{x=1}^{w} \bigoplus_{y=1}^{h} (\textbf{v}_{x} \otimes \textbf{v}_{y} \otimes \textbf{v}_{I_{bin}[x,y]})]$ \\
           &                          & Both        & 1D & \cite{Khaleghi2022} & $[\bigoplus_{x=1}^{(w*h)-n+1} (\textbf{v}_{x} \otimes \bigotimes_{j=0}^{n-1} (\rho^{j}\textbf{v}_{x[i+j]}))]$ \\ \cline{2-6}
           & Linear & & & \cite{Gallant2016, Kussul1992, Weiss2016} & $[\bigoplus_{x=1}^{w} \bigoplus_{y=1}^{h} (\textbf{v}_{x} \otimes \textbf{v}_{y} \otimes \textbf{v}_{I_{bin}[x,y]})]$ \\ \hline
Hybrid & \multicolumn{3}{r}{Feature extraction's output} & \cite{Karvonen2019, Kleyko2017a, Yilmaz2015, Zou2021b} & $\textbf{v}_{output}$ \\ \cline{2-6}
           & \multicolumn{3}{r}{Record-based feature encoding} & \cite{Kussul1991b} & $[\bigoplus_{i = 1}^{n} (\textbf{v}_{f[i]} \otimes \textbf{v}_{i})]$ \\ \hline
\end{tabular}
\end{table*}

\subsubsection{Native HDC} \label{sec:4.1.1}

Assume an image $I$ of size $w \times h$ is given as an input which is binarized, denoted here as $I_{bin}$. The binarized image is either flattened into an array $p$ of length $w*h$ where $p[x]$ is the value of the pixel in the array $p$ at position $x$, or used in its original 2D format where $I_{bin}[x,y]$ is the value of the pixel in the binary image $I_{bin}$ at position $(x,y)$. \\

\noindent The native HDC encoding methods can be further divided into two categories depending on whether position is encoded while preserving similarity between nearby positions (i.e., linearly mapped) or not (i.e., orthogonally mapped). \\

\noindent \textbf{(A) Orthogonally mapped position vectors}

\noindent \textit{(a) Permutation} \\
\noindent When considering the flattened image, a unique random HV is assigned to each pixel position in the array $p$ after which the obtained position HV $\textbf{v}_{x}$ is shifted with one position if the corresponding pixel value $p[x]$ is one and not shifted if it is zero \cite{Hassan2022, Kleyko2016, Kleyko2017b, Manabat2019}. To encode the 2D binarized image, two unique permutations $\rho_{X}$ and $\rho_{Y}$ are assigned to represent the x- and y-axis of the image, respectively. These permutations are applied $x$ and $y$ times, respectively, to the pixel value HV $\textbf{v}_{I_{bin}[x,y]}$ \cite{Kleyko2020, Kussul2006, Mitrokhin2019, Rachkovskij2022b}. \\

\noindent \textit{(b) Binding} \\
A unique random HV is assigned to each possible pixel value (i.e., zero and one). Thereafter, the pixel value HV $\textbf{v}_{p[x]}$ or $\textbf{v}_{I_{bin}[x,y]}$ is bound with its corresponding position HV $\textbf{v}_{x}$ or ($\textbf{v}_{x} \otimes \textbf{v}_{y}$) for the flattened image \cite{Bosch2022, Duan2022a, Ma2022, Ma2021, Watkinson2021, Yang2017b} or 2D image \cite{Kelly2013}, respectively, which are mapped orthogonally. \\

\noindent \textit{(c) Combination of permutation and binding} \\
In analogy to the $n$-gram encoding in language identification applications \cite{Rahimi2016b}, \textcite{Khaleghi2022} apply a sliding window of length $n$ to the image. The window is then encoded by binding all pixel value HV's which are permuted based on the position in the window, i.e., the first pixel value HV is not permuted, the second is permuted once, the third is twice permuted, etc. This could be seen as extracting local features from the image. To account for the global position of these features in the image, each window HV is bound with a random position HV. \\

\noindent The encoding approaches mentioned so far represent similar pixels at nearby positions by dissimilar HVs, because of the property of permutation that a permuted HV is dissimilar to its original, and because of orthogonal position HVs. Hence, these encoding approaches do not preserve similarity which might be crucial to solve an image classification task. \\

\noindent \textbf{(B) Linearly mapped position vectors} \label{sec:4.1.1.2}

\noindent \textcite{Gallant2016, Kussul1992, Weiss2016} apply \textit{linear mapping} such that nearby $x$ and $y$ positions are represented by similar HVs. The image is then encoded using the binding operation for a 2D image, as mentioned in section \ref{sec:4.1.1}(A,b). \\

\noindent An alternative approach to preserve similarity for nearby positions is proposed by \textcite{Frady2022, Komer2019, Voelker2021} who make use of fractional binding. For this, two random HVs $\textbf{x}$ and $\textbf{y}$ are assigned to represent the x- and y-axis, respectively. The $(x,y)$ position is then constructed as $\textbf{x}^{x} \otimes \textbf{y}^{y}$ where $\textbf{x}^{x} = \bigotimes_{n=1}^{x} \textbf{x}$, i.e., the HV $\textbf{x}$ is repeatedly bound with itself $x$ times. This bound pair representing the position is then bound with the pixel value HV $\textbf{v}_{I_{bin}[x,y]}$. However, this type of position encoding cannot be applied to binary HVs since all even (or odd) positions would be represented by the same HV.

\subsubsection{Hybrid HDC} \label{sec:4.1.2}

Instead of encoding the raw image by means of HD vector operations, external non-HD-based feature extraction methods are used. These approaches can be subdivided in two categories: \textbf{(a)} those that use the output layer of a neural network (NN) or cellular automata (CA) as single feature HV to represent the image \cite{Karvonen2019, Kleyko2017a, Yilmaz2015, Zou2021b}; and \textbf{(b)} those that use external methods (NN or other) to extract multiple features which are encoded via the record-based encoding (Figure \ref{fig:spatial}) \cite{Kussul1991b}. \\

\subsection{Proposed unified framework} \label{sec:4.2}

\begin{figure*}[!hbt]
\centering
\includegraphics[width=\linewidth]{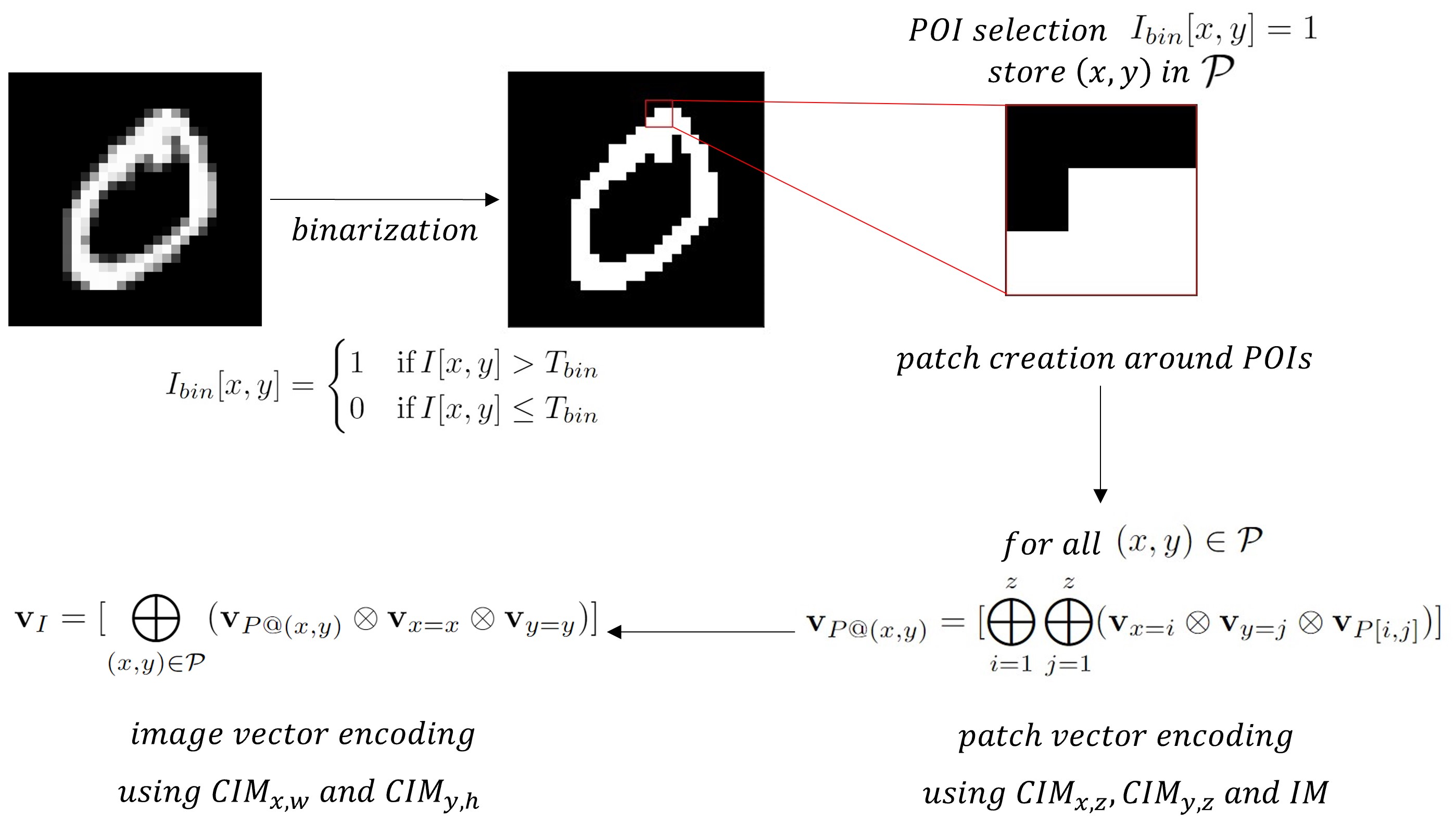}
\caption{Schematic overview of the proposed unified encoding framework for a training sample of the MNIST data set with size 28x28 using a patch size of 3x3 around the POIs ($z = 3$, $h = 28$ and $w = 28$).}
\label{fig:overviewencoding}
\end{figure*}

Figure \ref{fig:overviewencoding} gives an overview of the proposed approach to encode binarized images which can be divided into four steps: (1) binarization, (2) POI selection and patch creation around POIs, (3) patch vector encoding, and (4) image vector encoding. \\

\noindent \textbf{(1) Binarization.} As a first step, the pixel values of an input image $I$ are binarized using a predefined binarization threshold $T_{bin}$:
\begin{equation} \label{eq:binarization}
    I_{bin}[x,y] = 
    \begin{cases}
        1 & \text{if} \, I[x,y] > T_{bin} \\
        0 & \text{if} \, I[x,y] \leq T_{bin}
    \end{cases}
\end{equation} \\

\noindent \textbf{(2) POI selection and patch creation around POIs.} Point of interests (POIs) are selected as pixels with $I_{bin}[x,y] = 1$. Thereafter, a square patch $P$ of predefined size $z$ is drawn around each POI (in Figure \ref{fig:overviewencoding}, $z=3$). \\

\noindent \textbf{(3) Patch vector encoding.} Each pixel in the patch is encoded as the binding of three vectors: the HV representing its binary value $P[x,y]$ (stored in $IM$, one random vector for value 0 and another random vector for value 1), the HV corresponding to its x position in the patch and the one for the y position in the patch. The x and y position HVs are stored in two separate CIMs ($CIM_{x,z}$ and $CIM_{y,z}$), both containing $z$ vectors that are mapped with orthogonal mapping. The resulting patch vector for the POI with position (x,y) is then obtained by bundling all pixel vectors and binarizing the obtained bundle:
\begin{equation}
    \textbf{v}_{P@(x,y)} = [\bigoplus_{i=1}^{z} \bigoplus_{j=1}^{z} (\textbf{v}_{x=i} \otimes \textbf{v}_{y=j} \otimes \textbf{v}_{P[i,j]})]
\end{equation}
for all $(x,y) \in \mathcal{P}$. The encoding of patch vectors around POIs can be seen as extracting local features of the image in analogy to \textcite{Kussul2004}, \textcite{Kussul2006} and \textcite{Curtidor2021}, but here only native HD arithmetic operations are used instead of relying on an NN-based feature extractor. \\

\noindent \textbf{(4) Image vector encoding.} After obtaining the patch vectors of all POIs, each individual patch vector is bound with the HVs representing the corresponding POI's x and y position in the original image $I$ (stored in $CIM_{x,w}$ and $CIM_{y,h}$) to capture the global positional information of the extracted local features. The binarized bundling of all these patch vectors bound with its POI's position results in the image vector:
\begin{equation}
    \textbf{v}_{I} = [\bigoplus_{(x,y) \in \mathcal{P}} (\textbf{v}_{P@(x,y)} \otimes \textbf{v}_{x=x} \otimes \textbf{v}_{y=y})]
\end{equation}
The $CIM_{x,w}$ and $CIM_{y,h}$ are mapped with our proposed \textit{local linear mapping} instead of original \textit{linear mapping} (see section \ref{sec:3.3}) to capture small dependencies in position while ignoring large ones. \\

\section{Experiments} \label{sec:5}

The abovementioned proposed approach to encode binarized images is tested on two known, publicly available data sets: \\
(1) MNIST data set \cite{LeCun1998}. This data set includes 70,000 28x28 gray scale images of ten different handwritten digits. \\
(2) Fashion-MNIST data set \cite{Xiao2017}. This data set contains 7,000 28x28 gray scale images of fashion products for each of ten categories, i.e., 70,000 images in total. \\
Both data sets are split in a training set of 60,000 images (6,000 for each class) and a test set of 10,000 images (1,000 for each class). The pixel values range from 0 to 255. \\

\subsection{\textit{Local linear mapping}} \label{sec:5.1}

At first, the concept of \textit{local linear mapping} is tested using pixel-wise encoding on the whole image, without using POI encoding. The image is thus encoded as:
\begin{equation}
    \textbf{v}_{I} = [\bigoplus_{x=1}^{w} \bigoplus_{y=1}^{h} (\textbf{v}_{x} \otimes \textbf{v}_{y} \otimes \textbf{v}_{I_{bin}[x,y]})]
\end{equation}
The number of splits $S$ in the CIMs storing $\textbf{v}_{x}$ and $\textbf{v}_{y}$ is treated as a hyperparameter and tested to be equal to 1, 3, 5, 7, 9 and 28 of which the second from last is the maximal number of splits possible for a 28x28 image, since otherwise only 2 vectors would be in a particular split and thus will be orthogonal. Note again that using only 1 split corresponds to the traditional \textit{linear mapping} and will be treated as the baseline HDC framework, and using 28 splits corresponds to orthogonal mapping. The images are binarized following Equation \ref{eq:binarization} with the binarization threshold equal to zero (i.e., $T_{bin} = 0$). \\

\subsection{Proposed unified framework} \label{sec:5.2}

In the second part of the experiments, the \textit{local linear mapping} is applied together with the POI encoding. This encoding approach requires to determine the settings of two hyperparameters: the number of splits for \textit{local linear mapping} $S$ and the patch size $z$ around each POI. As will be presented in more detail in section \ref{sec:6.1}, an increase in performance is seen when number of splits $S$ is increased from 1 to 9 in the first part of the experiments (section \ref{sec:5.1}). Hence, only the cases $S = 1$, $S = 9$ and $S = 28$ will be tested in this second part of experiments since we believe that the same increasing trend between 1 and 9 will be seen as in the first part. The tested settings for the patch size are 3, 5 and 7. In summary, all possible combinations of the following settings of the two hyperparameters will be tested: $S = \{1,9,28\}$ and $z = \{3,5,7\}$. The images are again binarized following Equation \ref{eq:binarization} with the binarization threshold equal to zero (i.e., $T_{bin} = 0$). \\

\subsection{Evaluation} \label{sec:5.3}

The different combinations of settings are tested by means of 10-fold cross validation (CV) on the training set. This means that the 60,000 training images are split in ten parts. The algorithm is trained on 54,000 images and validated on the remaining 6,000 images which is repeated ten times while each time taking a different set of 6,000 validation images. The training procedure is performed iteratively for a maximum of 1,000 iterations while saving the classifier with the best accuracy. After every 100 iterations, it is evaluated whether this best training accuracy exceeds 99\% accuracy. If this is the case, the training procedure is terminated and the classifier with the best accuracy is used on the validation set. The performance of the HDC classifier for each combination of hyperparameter settings is documented as the average validation accuracy over the ten runs of the 10-fold CV. As such, the combination of hyperparameter settings yielding the largest average validation accuracy is selected and used to train on the entire training set (i.e., all 60,000 images) and tested on the 10,000 test images for ten independent runs. Finally, the average test accuracy over these ten independent runs is calculated. \\

\subsection{Robustness analysis} \label{sec:5.4}

To test the robustness to noise and blur of the proposed encoding approach, the MNIST-C data set which is proposed as a robustness benchmark for computer vision by \textcite{Mu2019} is used. This data set includes the 60,000 training and 10,000 test images of the original MNIST data set \cite{LeCun1998} to which several different corruptions are applied, including shot noise, impulse noise, glass blur, motion blur and spatter which are of particular interest in the current article to test noise and blur robustness. The HDC model with the proposed encoding is trained on the original 60,000 training images (i.e., without corruptions) with the baseline setting of hyperparameters ($S = 1$ and no POI selection) and the setting yielding the best validation accuracy after 10-fold CV. Both trained HDC classifiers are then tested on the five selected corrupted test sets of 10,000 images for which a test accuracy averaged over ten independent runs is calculated.

\section{Results} \label{sec:6}

\subsection{\textit{Local linear mapping}} \label{sec:6.1}

The results of the experiments testing the effect of the number of splits in \textit{local linear mapping} are presented in Table \ref{tab:resultspart1}. The table includes the accuracy on the training set, the accuracy on the validation set and the number of iterations needed to reach the maximal accuracy on the training set, averaged over the ten folds of the 10-fold CV for both the MNIST and Fashion-MNIST data set. As mentioned previously, the number of splits equal to 1 ($S = 1$) is treated as our baseline since this does not use \textit{local linear mapping} nor POI encoding. As such, the baseline average validation accuracy is 60.78\% for MNIST and 62.65\% for Fashion-MNIST. \\

\noindent An increase in performance is seen when increasing the number of splits used in \textit{local linear mapping} from 1 to 9. The largest validation accuracy is 93.21\% for MNIST for $S = 9$ and 80.98\% for Fashion-MNIST for $S = 28$, which is an increase of 32.43\% and 18.33\%, respectively. In the case of MNIST, the classifier with orthogonal mapping ($S = 28$) reaches an accuracy that lies inbetween the baseline and largest obtained accuracy, while this setting yields the highest accuracy for Fashion-MNIST. \\

\begin{table*}[!hbt]
\centering
\caption{Accuracy (\%) on the training and validation set and the number of iterations needed to reach the best training accuracy, averaged over the ten folds of 10-fold cross validation for the MNIST and Fashion-MNIST data set and for the different settings of the number of splits $S$ used in \textit{local linear mapping}. Data are \textit{mean} ($\pm$ \textit{standard deviation}) and in \textbf{bold} is the best validation accuracy for each data set.}
\label{tab:resultspart1}
\begin{tabular}{@{}c|ccc|ccc@{}}
\hline
\multicolumn{1}{c}{} & \multicolumn{3}{c}{\textbf{MNIST}} & \multicolumn{3}{c}{\textbf{Fashion-MNIST}} \\
\multirow{2}{*}{$S$} & Training & Validation & \multirow{2}{*}{Iteration} & Training & Validation & \multirow{2}{*}{Iteration} \\
 & Accuracy & Accuracy & & Accuracy & Accuracy & \\
\hline
1 & 61.87 ($\pm$ 1.04) & 60.78 ($\pm$ 1.57) & 590 ($\pm$ 233) & 62.80 ($\pm$ 1.20) & 62.65 ($\pm$ 1.71) & 390 ($\pm$ 300) \\
3 & 86.21 ($\pm$ 2.08) & 84.54 ($\pm$ 2.30) & 555 ($\pm$ 332) & 74.55 ($\pm$ 0.53) & 73.69 ($\pm$ 0.67) & 626 ($\pm$ 289) \\
5 & 93.74 ($\pm$ 0.67) & 91.89 ($\pm$ 0.99) & 219  ($\pm$ 95) & 78.25 ($\pm$ 0.79) & 76.63 ($\pm$ 0.90) & 606 ($\pm$ 248) \\
7 & 94.69 ($\pm$ 0.98) & 92.67 ($\pm$ 0.66) & 121  ($\pm$ 92) & 79.25 ($\pm$ 1.57) & 77.42 ($\pm$ 1.36) & 673 ($\pm$ 276)\\
9 & 95.68 ($\pm$ 0.77) & \textbf{93.21} ($\pm$ 0.61) & 191 ($\pm$ 102) & 79.74 ($\pm$ 1.35) & 77.92 ($\pm$ 1.35) & 380 ($\pm$ 250) \\
28 & 94.56 ($\pm$ 0.44) & 90.53 ($\pm$ 0.87) & 88 ($\pm$ 52) & 84.64 ($\pm$ 0.70) & \textbf{80.98} ($\pm$ 0.82) & 758 ($\pm$ 230) \\
\hline
\end{tabular}
\end{table*}

\subsection{Proposed unified framework} \label{sec:6.2}

Table \ref{tab:resultspart2} includes the results showing the effect of the hyperparameters (i.e., the number of splits $S$ in \textit{local linear mapping} and the patch size $z$ in POI encoding) for our proposed encoding approach. The table again includes the accuracy on the training set, the accuracy on the validation set and the number of iterations needed to reach the maximal accuracy on the training set, averaged over the ten folds of the 10-fold CV for both the MNIST and Fashion-MNIST data set. \\

\noindent Similar to the previous section, the validation accuracy is much larger in the case of $S = 9$ compared to $S = 1$, i.e., 96.48\% - 97.05\% versus 78.22\% - 91.33\% for MNIST and 84.30\% - 85.30\% versus 66.70\% - 77.54\% for Fashion-MNIST. This is also seen for $S = 28$ which shows accuracies in the range 93.27\% - 94.41\% for MNIST and 79.60\% - 83.98\% for Fashion-MNIST. An increase in performance is also seen with increasing patch size $z$. \\

\noindent The best achieved validation accuracy is 97.05\% for MNIST with $S = 9$ and $z = 7$ and 85.30\% for Fashion-MNIST with $S = 9$ and $z = 5$. This corresponds to an increase in performance of 36.27\% for MNIST and 22.65\% for Fashion-MNIST compared to their baseline accuracy ($S = 1$ in Table \ref{tab:resultspart1}). These settings for the two hyperparameters yielding the best validation accuracy are used to test the HDC classifier on the test set in the next section. \\

\begin{table*}[!hbt]
\centering
\caption{Accuracy (\%) on the training and validation set and the number of iterations needed to reach the best training accuracy, averaged over the ten folds of 10-fold cross validation for the MNIST and Fashion-MNIST data set and for the different settings of the number of splits $S$ used in \textit{local linear mapping} and of the patch size $z$ used in POI encoding. Data are \textit{mean} ($\pm$ \textit{standard deviation}) and in \textbf{bold} is the best validation accuracy for each data set.}
\label{tab:resultspart2}
\begin{tabular}{@{}cc|ccc|ccc@{}}
\hline
\multicolumn{1}{c}{} & \multicolumn{1}{c}{} & \multicolumn{3}{c}{\textbf{MNIST}} & \multicolumn{3}{c}{\textbf{Fashion-MNIST}} \\
\multirow{2}{*}{$S$} & \multirow{2}{*}{$z$} & Training & Validation & \multirow{2}{*}{Iteration} & Training & Validation & \multirow{2}{*}{Iteration} \\
 & & Accuracy & Accuracy & & Accuracy & Accuracy & \\
\hline
1  & 3 & 78.79 ($\pm$ 1.78) & 78.22 ($\pm$ 1.78) & 545 ($\pm$ 248) & 66.87 ($\pm$ 0.57) & 66.70 ($\pm$ 0.77) & 412 ($\pm$ 227) \\
1  & 5 & 87.81 ($\pm$ 1.22) & 87.12 ($\pm$ 1.75) & 614 ($\pm$ 292) & 74.49 ($\pm$ 0.93) & 74.07 ($\pm$ 1.35) & 861 ($\pm$ 69) \\
1  & 7 & 92.07 ($\pm$ 2.72) & 91.33 ($\pm$ 2.42) & 539 ($\pm$ 239) & 77.98 ($\pm$ 0.99) & 77.54 ($\pm$ 1.21) & 468 ($\pm$ 253) \\
9  & 3 & 98.52 ($\pm$ 0.38) & 96.48 ($\pm$ 0.38) & 123 ($\pm$ 38) & 85.71 ($\pm$ 0.93) & 84.30 ($\pm$ 0.89) & 777 ($\pm$ 223) \\
9  & 5 & 99.30 ($\pm$ 0.21) & 96.93 ($\pm$ 0.31) & 108 ($\pm$ 33) & 86.86 ($\pm$ 0.49) & \textbf{85.30} ($\pm$ 0.53) & 781 ($\pm$ 146) \\
9  & 7 & 99.27 ($\pm$ 0.08) & \textbf{97.05} ($\pm$ 0.40) & 95 ($\pm$ 7) & 86.69 ($\pm$ 0.65) & 84.96 ($\pm$ 0.84) & 799 ($\pm$ 200) \\
28 & 3 & 97.30 ($\pm$ 0.53) & 93.27 ($\pm$ 0.81) & 56 ($\pm$ 14) & 81.53 ($\pm$ 1.42) & 79.60 ($\pm$ 1.38) & 597 ($\pm$ 235) \\
28 & 5 & 99.43 ($\pm$ 0.29) & 94.00 ($\pm$ 0.68) & 97 ($\pm$ 32) & 85.52 ($\pm$ 0.85) & 82.88 ($\pm$ 0.69) & 579 ($\pm$ 277) \\
28 & 7 & 99.66 ($\pm$ 0.13) & 94.41 ($\pm$ 0.60) & 95 ($\pm$ 8) & 86.94 ($\pm$ 0.51) & 83.98 ($\pm$ 0.55) & 752 ($\pm$ 173) \\
\hline
\end{tabular}
\end{table*}

\subsection{Evaluation on the test set} \label{sec:6.3}

Table \ref{tab:resultspart3} shows the results obtained when setting the hyperparameters to the values yielding the best validation accuracy obtained in previous section. The table shows the accuracy on the entire training set, the accuracy on the unseen test set and the number of iterations needed to obtain the best training accuracy, averaged over ten independent runs. An average accuracy of 97.35\% is reached on the test set of MNIST. For the Fashion-MNIST data set, an average test accuracy of 84.12\% is obtained.

\begin{table*}[!hbt]
\centering
\caption{Accuracy (\%) on the full training and unseen test set and the number of iterations needed to reach the best training accuracy, averaged over ten independent runs for the MNIST ($S = 9$ and $z = 7$) and Fashion-MNIST ($S = 9$ and $z = 5$) data set. Data are \textit{mean} ($\pm$ \textit{standard deviation}).}
\label{tab:resultspart3}
\begin{tabular}{@{}ccc|ccc@{}}
\hline
\multicolumn{3}{c}{\textbf{MNIST}} & \multicolumn{3}{c}{\textbf{Fashion-MNIST}} \\
Training & Test & \multirow{2}{*}{Iteration} & Training & Test & \multirow{2}{*}{Iteration} \\
Accuracy & Accuracy & & Accuracy & Accuracy & \\
\hline
99.37 ($\pm$ 0.33) & \textbf{97.35} ($\pm$ 0.12) & 126 ($\pm$ 46) & 86.55 ($\pm$ 0.57) & \textbf{84.12} ($\pm$ 0.56) & 808 ($\pm$ 220) \\
\hline
\end{tabular}
\end{table*}

\subsection{Robustness analysis} \label{sec:6.4}

Table \ref{tab:resultspart4} contains the results obtained during the analysis of robustness to noise and blur. The table includes the accuracy on the original and five selected corrupted test sets, averaged over ten independent runs for the MNIST-C data set with the hyperparameters set to the baseline setting ($S = 1$ and no POI selection) and the setting yielding the best validation accuracy with 10-fold CV ($S = 9$ and $z = 7$). The last row of the table contains the average test accuracy across all five corrupted test sets. As such, it is seen that the best hyperparameters setting achieves an average test accuracy of 72.58\%, which is an increase of 39.16\% compared to the baseline setting which achieves 33.42\% average test accuracy.

\begin{table*}[!hbt]
\centering
\caption{Accuracy (\%) on the original and five selected corrupted test sets, averaged over ten independent runs for the MNIST-C data set with the baseline hyperparameters ($S = 1$ and no POI selection) and the best hyperparameters ($S = 9$ and $z = 7$). The last row contains the average test accuracy across all five corrupted test sets for each setting of hyperparameters. Data are \textit{mean} ($\pm$ \textit{standard deviation}).}
\label{tab:resultspart4}
\begin{tabular}{@{}ccc@{}}
\hline
\multicolumn{3}{c}{\textbf{MNIST-C}}\\
\textbf{Corruption} & \textbf{Baseline hyperparameters} & \textbf{Best hyperparameters} \\
\hline
None          & 62.21 ($\pm$ 0.81) & 97.35 ($\pm$ 0.12) \\
\hline
Shot Noise    & 41.98 ($\pm$ 7.90) & 96.04 ($\pm$ 0.17) \\
Impulse Noise & 57.60 ($\pm$ 1.18) & 91.41 ($\pm$ 0.58) \\
Glass Blur    & 18.87 ($\pm$ 8.61) & 55.84 ($\pm$ 1.54) \\
Motion Blur   & 11.76 ($\pm$ 1.31) & 38.46 ($\pm$ 2.63) \\
Spatter       & 36.91 ($\pm$ 4.38) & 81.13 ($\pm$ 0.49) \\
\hline
Average       & 33.42 ($\pm$ 16.44) & 72.58 ($\pm$ 22.01) \\
\end{tabular}
\end{table*}

\section{Discussion} \label{sec:7}

\subsection{Analysis of results} \label{sec:7.1}

The results in Table \ref{tab:resultspart1} for pixel-wise encoding show that the proposed \textit{local linear mapping} for position encoding outperforms \textit{linear mapping}. More specifically, there is an increase in performance with increasing number of splits used in \textit{local linear mapping}. This interesting finding indicates the importance of discriminating better smaller differences in position in the image instead of large differences. This is a result of the splits in \textit{local linear mapping} that represents two positions that are far apart with orthogonal HVs, and only HVs of close positions are similar. By contrast, in \textit{linear mapping}, the HVs of both close and far positions have a certain degree of similarity. \\

\noindent Another finding that stands out from the results reported earlier is a remarkable increase in performance when encoding patches around POIs (Table \ref{tab:resultspart2}) compared to pixel-wise encoding (Table \ref{tab:resultspart1}). Several factors could explain this observation. Firstly, background pixels are ignored with POI encoding, limiting unnecessary information. Secondly, local features are extracted around each POI such that the local neighborhood of each POI is taken into account. \\

\noindent In addition, employing \textit{local linear mapping} to encode the global position of POIs in the image improves the performance compared to using \textit{linear mapping} (Table \ref{tab:resultspart2}). This finding is in line with the results obtained in Table \ref{tab:resultspart1} and can be explained in a similar way as done above. \\

\noindent Finally, the results of the robustness analysis indicate that the proposed encoding approach after hyperparameter selection shows a higher robustness to noise and blur than the baseline HDC encoding approach (Table \ref{tab:resultspart4} and Section \ref{sec:7.3}).

\subsection{Comparison to the  state-of-the-art} \label{sec:7.2}

\subsubsection{MNIST data set} \label{sec:7.2.1}

Table \ref{tab:literatureMNIST} provides a summary for the comparison of our obtained result for MNIST (i.e., 97.35\%) with other studies found in the literature. \\

\noindent The proposed approach of POI encoding with \textit{local linear mapping} outperforms all methods categorized in Native HDC. This includes the methods applying the permutation operation to encode position of pixels in the flattened image (section \ref{sec:4.1.1}(A,a)), i.e., \textcite{Manabat2019} and \textcite{Hassan2022} report an accuracy of 79.87\% and 86\%, respectively. \\

\noindent Our obtained result for MNIST is also better compared to several studies using the binding operation for position encoding in the flattened image (section \ref{sec:4.1.1}(A,b)). Namely, \textcite{Kazemi2021}, \textcite{Chang2021}, \textcite{Duan2022b}, \textcite{Bosch2022}, \textcite{Chuang2020}, \textcite{Duan2022a}, \textcite{Ma2022} and \textcite{Zou2021a} report a baseline accuracy of 85\%, 87\%, 87.38\%, 88.3\%, 88.92\%, 89.28\%, 90.93\% and 92\%, respectively. \\

\noindent In addition, the $n$-gram-based encoding method to extract local features by \textcite{Khaleghi2022} reaches an accuracy of 94.0\% which we outperform by using \textit{local linear mapping} instead of orthogonal mapping to encode global positional information. \\

\noindent \textcite{Hernandez2021a} propose OnlineHD that is able to increase their baseline performance of 91\% to 97.5\%, which is slightly higher than our obtained accuracy. In OnlineHD, the baseline HDC training procedure is extended by updating the HDC model depending on how similar a sample is to the existing model. As such, the training procedure becomes more complex due to floating-point multiplications. \\

\noindent Other studies use the HDC framework in combination with additional non-HD methods (Hybrid HDC, section \ref{sec:4.1.2}), such as elementary CA which is used to derive the high-dimensional vector by \textcite{Karvonen2019} resulting in an accuracy of 74.06\%. \textcite{Zou2021b} extracts low-level features with an SNN before using HDC reaching an accuracy of 90.5\%. \textcite{Duan2022a} maps the HDC framework to an equivalent BNN where the class HVs in binary HDC are reflected by the trained binary weights in BNN reaching an accuracy of 94.74\%. Random Fourier Features (RFF) are used by \textcite{Yu2022} during the encoding of the images resulting in 95.4\% accuracy. \textcite{Duan2022b} and \textcite{Ma2022} derive an HDC model from an NN which results in 92.72\% and 96.71\% accuracy, respectively. Our proposed encoding approach using only native HD vector operations outperforms these hybrid HDC methods. \\

\noindent Nevertheless, more advanced hybrid HDC methods obtain better results. \textcite{Kussul2004} and \textcite{Kussul2006} reach a higher accuracy of 99.2\% and 99.5\% with the NN-based local feature extraction. \textcite{Zou2021a} report an accuracy of 97.5\% by extending the HDC encoding framework with manifold learning. \textcite{Rachkovskij2022b} extracts local binary pattern (LBP) features, proposes a shift-equivariant similarity-preserving scheme for position encoding and uses a large margin perceptron for classification reaching an accuracy of 98.5\% with a vector dimension of 10,000. \textcite{Liang2022} and \textcite{Poduval2021} extract features from the original images and apply record-based encoding to obtain a performance of 94.8\% and 99\%, respectively. \\

\noindent Finally, several works increase the complexity of HDC by using multi-bit representations instead of single-bit (i.e., binary). \textcite{Kazemi2021} use 3-bit precision to increase the information representation capacity of HVs and achieve an accuracy of 95.5\%. \textcite{Yu2022} use HVs that have more complex elements than binary achieving 96.6\% accuracy. The fully integer model by \textcite{Chuang2020} reaches a slightly higher accuracy of 98.09\%. With only the latter achieving a slightly higher accuracy than ours, we can conclude that our proposed binary, native HDC method using \textit{local linear mapping} and POI encoding achieves comparable results with these more complex multi-bit HDC methods.

\begin{table*}[!hbt]
\centering
\caption{Obtained accuracies (\%) found in the literature for the MNIST data set. In \textbf{bold} is the obtained accuracy with the proposed encoding approach.}
\label{tab:literatureMNIST}
\begin{tabular}{@{}rrrl@{}}
\hline
\textbf{Category} & \textbf{Method} & \textbf{Accuracy} & \textbf{Reference} \\
\hline
Native HDC & Permutation 1D               & 79.87 & \textcite{Manabat2019} \\
           &                              & 86    & \textcite{Hassan2022} \\
           & Binding 1D                   & 85    & \textcite{Kazemi2021} \\
           &                              & 87    & \textcite{Chang2021} \\
           &                              & 87.38 & \textcite{Duan2022b} \\
           &                              & 88.3  & \textcite{Bosch2022} \\
           &                              & 88.92 & \textcite{Chuang2020} \\
           &                              & 89.28 & \textcite{Duan2022a} \\
           &                              & 90.93 & \textcite{Ma2022} \\
           &                              & 91    & \textcite{Hernandez2021a} \\
           &                              & 92    & \textcite{Zou2021a} \\
           & Permutation \& Binding 1D    & 94.0  & \textcite{Khaleghi2022} \\
           & \textbf{\textit{Local linear mapping} and POI} & \textbf{97.35} & \textbf{Ours} \\
\hline
HDC with adaptive training & Binding 1D   & 97.5  & \textcite{Hernandez2021a} \\
\hline
Hybrid HDC & Elementary Cellular Automata & 74.06 & \textcite{Karvonen2019} \\
           & Spiking Neural Network       & 90.5  & \textcite{Zou2021b} \\
           & Binary Neural Network        & 94.74 & \textcite{Duan2022a} \\
           & Random Fourier Features      & 95.4  & \textcite{Yu2022} \\
           & Neural Network               & 92.72 & \textcite{Duan2022b} \\
           &                              & 94.8  & \textcite{Liang2022} \\
           &                              & 96.71 & \textcite{Ma2022} \\
           &                              & 99    & \textcite{Poduval2021} \\
           &                              & 99.2  & \textcite{Kussul2004} \\
           &                              & 99.5  & \textcite{Kussul2006} \\
           & Manifold learning            & 97.5  & \textcite{Zou2021a} \\
           & Local Binary Pattern         & 98.5  & \textcite{Rachkovskij2022b} \\
\hline
Multi-bit HDC &                           & 95.5  & \textcite{Kazemi2021} \\
              &                           & 96.6  & \textcite{Yu2022} \\
              &                           & 98.09 & \textcite{Chuang2020} \\
\hline
\end{tabular}
\end{table*}

\subsubsection{Fashion-MNIST data set} \label{sec:7.2.2}

Table \ref{tab:literatureFashionMNIST} provides a summary for the comparison of our obtained result for Fashion-MNIST (i.e., 84.12\%) with other studies found in the literature. \\

\noindent There are not as many studies available for the Fashion-MNIST data set as for MNIST. \textcite{Duan2022b} and \textcite{Duan2022a} report an accuracy of 79.24\% and 80.26\% for baseline HDC. Using hybrid HDC methods, \textcite{Yu2022} report an accuracy of 84.0\% when using RFF and reach 87.4\% using more complex elements in the HVs. \textcite{Duan2022b} and \textcite{Duan2022a} reach a higher accuracy of 85.47\% and 87.11\% by mapping the HDC model to an equivalent (B)NN. We can conclude that our proposed HDC method outperforms the native HDC methods but achieves a lower accuracy than the hybrid and multi-bit HDC methods.

\begin{table*}[!hbt]
\centering
\caption{Obtained accuracies (\%) found in the literature for the Fashion-MNIST data set. In \textbf{bold} is the obtained accuracy with the proposed encoding approach.}
\label{tab:literatureFashionMNIST}
\begin{tabular}{@{}rrrl@{}}
\hline
\textbf{Category} & \textbf{Method} & \textbf{Accuracy} & \textbf{Reference} \\
\hline
Native HDC & Binding 1D              & 79.24 & \textcite{Duan2022b} \\
           &                         & 80.26 & \textcite{Duan2022a} \\
           & \textbf{\textit{Local linear mapping} and POI} & \textbf{84.12} & \textbf{Ours} \\
\hline
Hybrid HDC & Random Fourier Features & 84.0  & \textcite{Yu2022} \\
           & Neural Network          & 85.47 & \textcite{Duan2022b} \\
           & Binary Neural Network   & 87.11 & \textcite{Duan2022a} \\
\hline
Multi-bit HDC &                      & 87.4 & \textcite{Yu2022} \\
\hline
\end{tabular}
\end{table*}

\subsection{Robustness analysis} \label{sec:7.3}

After selecting the hyperparameters yielding the best validation accuracy with 10-fold CV, the proposed encoding approach is more robust to images corrupted with noise and blur compared to the baseline encoding approach (Table \ref{tab:resultspart4}). Especially for the shot noise and impulse noise corruption, the average test accuracy is fairly equivalent to the average test accuracy achieved on non-corrupted images. For spatter, the average test accuracy is slightly dropped but the proposed approach is still able to identify around 81\% of the test images accurately. The average test accuracy drops the most for the glass blur and motion blur corruption where the proposed approach is able to classify respectively 55.84\% and 38.46\% of the images correctly. Still, this is an improvement of 36.97\% for glass blur and 26.7\% for motion blur compared to the baseline HDC encoding approach such that it can be concluded that the HDC classifier with our proposed encoding approach after hyperparameter selection has a high robustness to noise and blur with an average accuracy of 72.58\% across five different corrupted test sets. \\

\subsection{Future research} \label{sec:7.4}

As future work, we envisage evaluating and extending the proposed encoding approach for application to gray scale and color images, investigating the use of hierarchical (multi-layer) patches with HDC encoding and further extensions of the \textit{local linear mapping} concept for position encoding. \\

\noindent It could also be analysed how the HDC framework can be made even more robust to noise and corruptions such as glass blur and motion blur. \\

\section{Conclusion} \label{sec:8}

A novel light-weight approach to encode binarized images that preserves similarity of patterns at nearby locations while relying only on native HD arithmetic vector operations, and not making use of external methods for feature extraction, is introduced. The approach uses point of interest selection to derive local features of the image and \textit{local linear mapping} to encode the location of these local features in the image. After selecting the best settings for the four introduced hyperparameters with 10-fold cross validation, an accuracy of 97.35\% is reached on the test set for the MNIST data set and 84.12\% for the Fashion-MNIST data set. These results outperform other studies using baseline HDC with different encoding approaches and are on par with more complex hybrid HDC models. The proposed encoding approach also shows a higher robustness to noise and blur compared to the baseline encoding.

\end{multicols}

\section*{Author Contributions}

\textbf{Laura Smets:} Conceptualization, Methodology, Software, Writing - Original Draft, Visualization \textbf{Werner Van Leekwijck:} Conceptualization, Methodology, Writing - Review \& Editing \textbf{Ing Jyh Tsang:} Writing - Review \& Editing \textbf{Steven Latr\'e:} Writing - Review \& Editing, Supervision, Funding Acquisition

\section*{Declaration of Competing Interest}

The authors declare that they have no known competing financial interests or personal relationships that could have appeared to influence the work reported in this article.

\section*{Funding statement}

This work was supported by the Flemish Government under the "Onderzoeksprogramma Artifici\"ele Intelligentie (AI) Vlaanderen" programme.

\section*{Appendix: List of symbols}
\setcounter{table}{0}
\renewcommand{\thetable}{A\arabic{table}}

A summary of notation can be found in Table \ref{tab:symbols}.

\begin{table}[!hbt]
\centering
\caption{List of symbols. (HD = hyperdimensional, (C)IM = (continuous) item memory)}
\label{tab:symbols}
\begin{tabular}{@{}rl|rl@{}}
\hline
 \textbf{Symbol} & \textbf{Definition} & \textbf{Symbol} & \textbf{Definition} \\
\hline
$f$           & feature vector in input space   & $L$       & number of levels in ordinal/discrete data  \\
$n$           & number of bundled elements      & $j$       & $1...n$                                    \\
$D$           & HD vector dimension             & $d$       & $1...D$                                    \\
$s$           & similarity                      & $h$       & Hamming distance                           \\
 & & & \\
$\textbf{v}$ & vector in HD space $\mathcal{H}$ & $\textbf{B}$ & bundle in HD space $\mathcal{B}$ \\
$\textbf{s}$ & sample vector                    & $\textbf{S}$ & sample bundle                    \\
 & & & \\
$\mathcal{H}$ & vector HD space, $\{0,1\}^{D}$  & $\mathcal{B}$ & bundle HD space, $\mathbb{N}^{D}$ \\
$\oplus$      & bundling operator               & $[.]$         & majority rule \\
$\otimes$     & binding operator                & $\rho$        & permutation operator \\ 
 & & & \\
$I$       & input image                                       & $T_{bin}$ & binarization threshold      \\
$I_{bin}$ & binarized image $I$                               & $p$    & flattened image $I_{bin}$               \\
$w$      & width of image $I$                                 & $h$    & height of image $I$                            \\
$P$      & patch of $I_{bin}$                                 & $z$    & patch size                                     \\
$I_{bin}[x,y]$ & value of pixel at position $(x,y)$ in $I_{bin}$ & $p[x]$ & value of pixel at position $x$ in $p$ \\
$P[x,y]$   & value of pixel at position $(x,y)$ in patch $P$  & $S$    & number of splits in \textit{local linear mapping} \\
$\rho^{i}$ & permutation applied $i$ times                    & $M$ & type of mapping in patch \\
$O$ & orthogonal mapping in patch                             & $L$ & linear mapping in patch \\
$\rho_{X}$ & unique permutation for x-axis in $I$             & $\rho_{Y}$ & unique permutation for y-axis in $I$       \\
$\textbf{v}_{I_{bin}[x,y]}$ & HV representing pixel value $I_{bin}[x,y]$  & $\textbf{v}_{p[x]}$      & HV representing pixel value $p[x]$ \\
$\textbf{v}_{x}$ & HV representing position $x$ in $p$ or $I$ & $\textbf{v}_{y}$         & HV representing position $y$ in $I$ \\
$\textbf{x}$     & unique random HV for x-axis in $I$         & $\textbf{y}$             & unique random HV for y-axis in $I$ \\
$\textbf{v}_{output}$ & HV of (last) layer of a neural network & $n$                      & number of features \\
$\textbf{v}_{i}$         & HV representing the $i$th feature      & $\textbf{v}_{f[i]}$      & HV representing the value of the $i$th feature \\
$\mathcal{P}$ & set of (x,y) positions of POIs & $IM$ & IM storing binary pixel values \\
$CIM_{x,z}$ & CIM storing position vectors for $x$-axis & $CIM_{y,z}$ & CIM storing position vectors for $y$-axis \\
 & in patch $P$ with size $z$ & & in patch $P$ with size $z$ \\
$CIM_{x,w}$ & CIM storing position vectors for $x$-axis & $CIM_{y,h}$ & CIM storing position vectors for $y$-axis \\
 & in image $I$ with width $w$ & & in image $I$ with height $h$ \\
\hline
\end{tabular}
\end{table}

\printbibliography

\end{document}